\documentclass[journal]{IEEEtran}

\usepackage{times}
\usepackage{epsfig}
\usepackage{url}
\usepackage{graphicx}
\usepackage{amsmath}
\usepackage{amssymb}
\usepackage{mathtools, nccmath}
\usepackage[linesnumbered,ruled,vlined]{algorithm2e}
\usepackage{comment,color,soul}
\usepackage{color, colortbl}
\usepackage{makecell}
\setcellgapes{1.5pt}
\DeclarePairedDelimiter{\nint}\lfloor\rceil

\definecolor{LightCyan}{rgb}{0.88,1,1}
\definecolor{Gray}{gray}{0.9}

\ifCLASSINFOpdf
\else
\fi
\begin{document}
%
\title{HYPER-SNN: Towards Energy-efficient Quantized Deep Spiking Neural Networks for Hyperspectral Image Classification}
%
%
%

\author{Gourav~Datta,~\IEEEmembership{Student Member,~IEEE,}
        Souvik~Kundu,~\IEEEmembership{Student Member,~IEEE,}
        Akhilesh~R.~Jaiswal,~\IEEEmembership{Member,~IEEE,}
        Peter~A.~Beerel ~\IEEEmembership{Senior Member,~IEEE} 
\thanks{G. Datta, S. Kundu, A. R. Jaiswal and P. A. Beerel are with the Department of Electrical and Computer Engineering, University of Southern California, Los Angeles,
CA, 90089 USA e-mail: \{gdatta, souvikku, akhilesh, pabeerel\}@usc.edu.}
}


%
%

\markboth{HYPER-SNN: Towards Energy-efficient Quantized Deep Spiking Neural Networks for Hyperspectral Image Classification}%
{Shell \MakeLowercase{\textit{et al.}}: Bare Demo of IEEEtran.cls for IEEE Journals}
%



\maketitle

\begin{abstract}
Hyperspectral images (HSIs) provide rich spectral–spatial information across a series of contiguous spectral bands. However, the accurate processing of the spectral and spatial correlation between the bands requires the use of energy-expensive 3-D Convolutional Neural Networks (CNNs). To address this challenge, we propose the use of Spiking Neural Networks (SNNs) that are generated from iso-architecture CNNs and trained with quantization-aware gradient descent to optimize their weights, membrane leak, and firing thresholds.
During both training and inference, the analog pixel values of a HSI are directly applied to the input layer of the SNN without the need to convert to a spike-train. The reduced latency of our training technique combined with high activation sparsity yields significant improvements in computational efficiency. 
We evaluate our proposal using three HSI datasets on a 3-D and a 3-D/2-D hybrid convolutional architecture. We achieve overall accuracy, average accuracy, and kappa coefficient of $98.68\%$, $98.34\%$, and $98.20\%$ respectively with $5$ time steps (inference latency) and $6$-bit weight quantization on the Indian Pines dataset. In particular, our models achieved accuracies similar to state-of-the-art (SOTA) with ${\sim}560.6\times$ and ${\sim}44.8\times$ less compute energy on average over three HSI datasets than an iso-architecture full-precision and 6-bit quantized CNN, respectively. 

\end{abstract}


%
\IEEEpeerreviewmaketitle

\section{Introduction}
%
%
%
%
\label{sec:intro}

\IEEEPARstart{H}{yperspectral} imaging, which extracts rich spatial-spectral information about the ground surface, has shown immense promise in remote sensing \cite{chen2014survey}. It is currently used in several applications ranging from geological surveys \cite{wan2021geological}, to the detection of camouflaged vehicles \cite{papp2020detection}. In hyperspectral images (HSIs), each pixel can be considered as a high-dimensional vector where each entry corresponds to the spectral reflectivity \cite{chen2014survey} of a particular wavelength.  The goal of the classification task is to assign a unique semantic label to each pixel \cite{zheng2020fpga}.

For HSI classification, several spectral feature-based methods have been proposed, including support vector machine \cite{melgani2018svm}, random forest \cite{pal2003rf}, canonical correlation forest \cite{xia2017ccf}, and multinomial logistic regression \cite{krishnapuram2005mlr}. To improve the accuracy of HSI classification, researchers have integrated spatial features into existing learning methods \cite{campvalls2006hybrid}. Some spectral-spatial methods for classifying HSIs include fusing correlation coefficient and sparse representation \cite{tu2018fuse}, Boltzmann entropy-based band selection \cite{gao2019boltzmann}, joint sparse model and discontinuity preserving relaxation \cite{gao2018sparse}, and extended morphological profiles \cite{benediktsson2005em,li2013kernel}. Some of these methods have also been proposed to exploit the spatial context with various morphological operations for HSI classification. However, these spectral-spatial feature extraction methods rely on hand-designed descriptions, prior information, and empirical hyperparameters \cite{chen2014survey}.

Lately, convolutional neural networks (CNNs) have yielded higher accuracy than some hand-designed features \cite{krizhevsky2012imagenet}. CNNs have shown promise in multiple applications where visual information processing is required, including image classification \cite{he2016deep}, object detection \cite{shaoqing2017rcnn}, semantic segmentation \cite{he2018mask}, and depth estimation \cite{repala2019dual}. In particular, CNN-based methods act as an end-to-end feature extractor that consists of a series of hierarchical filtering layers for global optimization. The 2-D CNN stacked autoencoder \cite{chen2014survey} was the first attempt to extract deep features from its compressed latent space to classify HSIs. Following this work, \cite{makantasis2015hsi} employed a 2-D CNN model to extract the spatial information in a supervised manner and classify the raw hyperspectral images. The multibranch selective kernel network with attention \cite{fard2020kernel} and pixel-block pair based data augmentation techniques \cite{song2018feature} were developed to address the gradient vanishing and overfitting problems. To extract the spatial-spectral features jointly from the raw HSI, researchers proposed a 3-D CNN architecture \cite{hamida2018-3d}, which achieves even better classification results. Authors in \cite{lee2017hybrid,roy2020hybrid,luo2018hsi} created multiscale spatiospectral relationships using 3-D CNN and fused the features using a 2-D CNN to extract more robust representation of spectral–spatial information.   

However, the performance and success of multi-layer CNNs are generally associated with high power and energy costs \cite{li2016cost}. A typical hyperspectral image cube consists of several hundred spectral frequency bands, and hence, classifying these images using traditional CNNs require a large amount of computational power, especially when real time processing is necessary, as in target tracking or identification \cite{van2010tracking}. The high energy cost and the demand for deployment of HSI sensors in battery-powered edge devices motivates exploring alternative lightweight energy-efficient HSI classification models.

In particular, low-latency spiking neural networks (SNNs) \cite{pfeiffer2018snn} have gained attention 
because they can be more  
computational efficient than CNNs for a variety of applications, including image analysis. To achieve this goal, 
analog inputs are first encoded into a sequence of spikes using one of a variety of proposed encoding methods, including rate coding \cite{diehl2016conversion,dsnn_conversion_abhronilfin}, direct coding \cite{rathi2020diet}, temporal coding \cite{comsa_2020}, rank-order coding \cite{Kheradpisheh_2020},
phase coding \cite{kim_2018}, and other exotic coding schemes \cite{ammar_2019, datta2021training}. Among these, rate and direct coding have shown competitive performance on complex tasks \cite{diehl2016conversion,dsnn_conversion_abhronilfin} while others are either limited to simpler tasks such as learning the XOR function and classifying MNIST images or require a large number of spikes for inference. 

In particular, for rate coding, the analog value is converted to a spike train using a Poisson generator function with a rate proportional to the input pixel value. 
The number of timesteps $T$ in each train is inversely proportional to the quantization error in the representation, as illustrated in Fig. \ref{fig:if_model_in_image}(b). 
\cite{dsnn_conversion_abhronilfin}. In contrast, in direct-input encoding, the analog pixel values are fed into the first convolutional layer as multi-bit values that are fixed for all $T$ timesteps.
\cite{rathi2020diet}.

In addition to accommodating various forms of encoding inputs, supervised learning algorithms for SNNs have overcome various roadblocks associated with the discontinuous derivative of the spike activation function \cite{lee_dsnn,wu2019direct}. 
In particular, recent works have shown that SNNs can be efficiently converted from artifical neural networks (ANNs) by approximating the activation value of ReLU neurons with the firing rate of spiking neurons \cite{dsnn_conversion_abhronilfin}. Low-latency SNNs trained using ANN-SNN conversion, coupled with supervised training, have been able to perform at par with ANNs in terms of classification accuracy in traditional image classification tasks \cite{rathi2020diet}. This motivates this work which explores the effectiveness of SNNs for HSI classification. 


More specifically, this paper provides the following contributions:
\begin{itemize}
    \item We propose two convolutional architectures for HSI classification that can yield classification accuracies similar to state-of-the-art (SOTA) and are compatible with our ANN-SNN conversion framework.
    \item We propose a hybrid training algorithm that first converts an ANN for HSI classification to an iso-architecture SNN, and then trains the latter using a novel quantization-aware spike timing dependent backpropagation (Q-STDB) algorithm.
    \item We evaluate and compare the energy-efficiency of the SNNs obtained by our training framework, with standard ANNs, using appropriate energy models, which reveal that our SNNs trained for HSI classification can offer significant improvement in compute efficiency. 
\end{itemize}

The remainder of this paper is structured as follows. In Section \ref{sec:back} we present necessary background and related work. Section \ref{sec:snn_training} and \ref{sec:arch} describe our quantization-aware SNN training method and network architectures respectively. 
We present the detailed experimental evaluations of our proposal in Section \ref{sec:expt}. We show the improvement in energy-efficiency of our proposed SNN for all the HSI classification tasks in Section \ref{sec:edp_analysis}. Finally, the paper concludes in Section \ref{sec:concl}.

\section{Background and Related Work}\label{sec:back}
\subsection{SNN Modeling}

An SNN consists of a network of neurons that communicate via a sequence of spikes modulated by synaptic weights. 
The activity of pre-synaptic neurons modulates the membrane potential of postsynaptic neurons, generating an action potential or spike when the membrane potential crosses a firing threshold.
The spiking dynamics of a neuron are generally modeled using either the Integrate-and-Fire (IF) \cite{lu2020exploring} or Leaky-Integrate-and-Fire (LIF) model \cite{leefin2020}. Both IF and LIF neurons accumulate the input current into their respective states or membrane potentials. The  difference between the two models is that the membrane potential of a IF neuron does not change during the time period between successive input spikes while the LIF neuronal membrane potential leaks at a constant rate.
In this work, we use the LIF model to convert ANNs trained with ReLU activations, to SNNs, because the 
leak term provides a tunable control knob that can reduce inference latency and spiking activity.
The IF model can be characterized by the following differential equation 
\begin{figure}[t!]
\begin{center}
\includegraphics[width=0.45\textwidth]{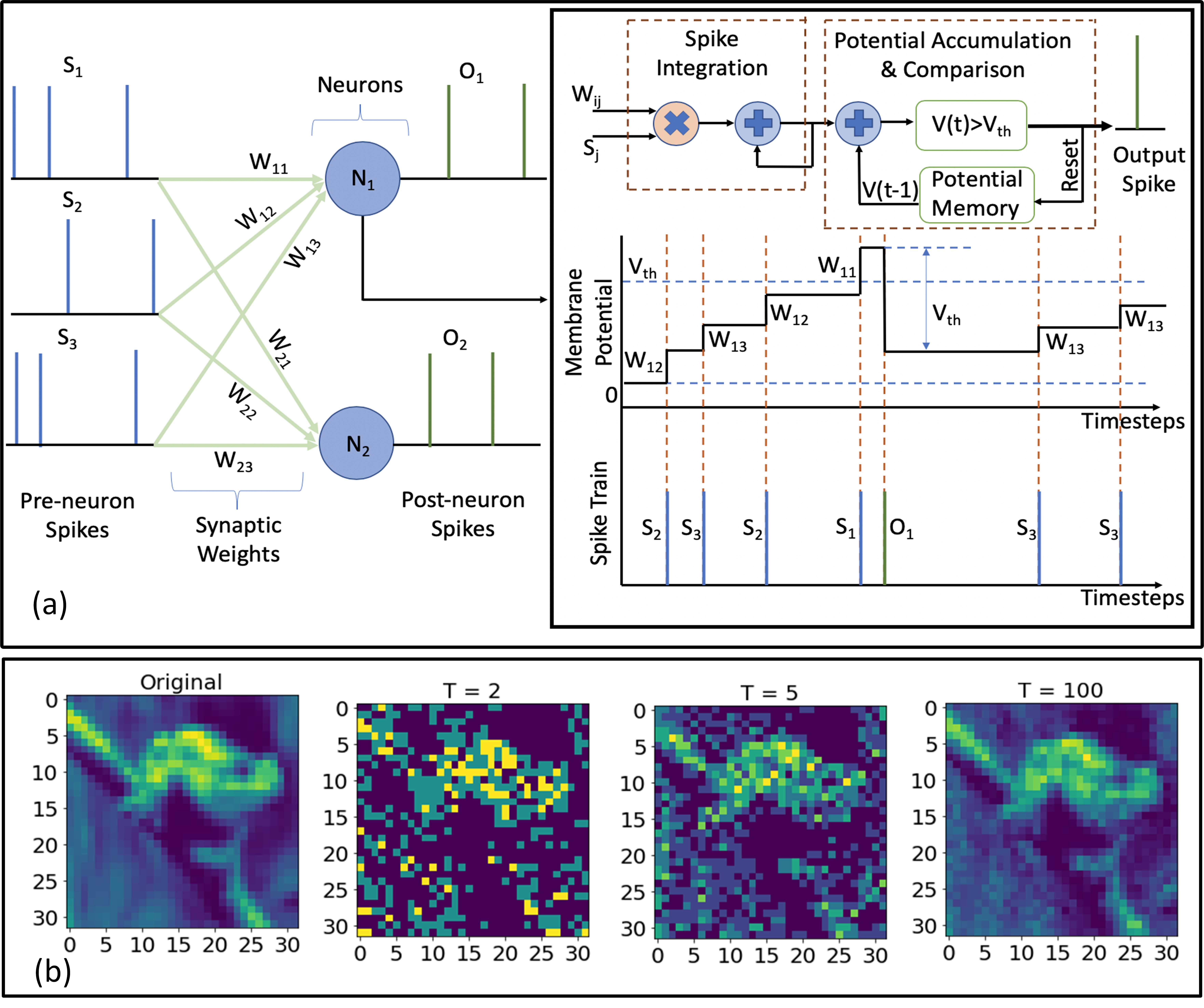}
\end{center}
\caption{(a) Feedforward fully-connected SNN architecture with integrate and fire (IF) spiking dynamics, (b) The spike input generated over several timesteps through a Poisson generator. It is clear that the larger the number of timesteps, the better the accumulated input spikes approximates the original input image.}
\label{fig:if_model_in_image}
\vspace{-5mm}
\end{figure}

\begin{equation}
C\frac{dU_i(t)}{dt}=I_i(t)=\sum_{j} W_{ij}\cdot{S_j(t)}
\label{eq:IF_continuous}
\end{equation}
\noindent
where $C$ is the membrane capacitance, $U_i(t)$ and $I_i(t)$ are the membrane potential and input synaptic current of the $i^{th}$ neuron at time $t$. As illustrated in Fig. \ref{fig:if_model_in_image}(a), $U_i(t)$ integrates the incoming (pre-neuron) binary spikes $S_j(t)$ multiplied by weights $W_{ij}$. The neuron generates an output spike when $U_i$ exceeds the firing threshold $V$. However, because of its continuous-time representation, Eq. \ref{eq:IF_continuous} is incompatible for implementation in common Machine Learning (ML) frameworks (e.g. Pytorch). Hence, we follow an iterative version 
evaluated in discrete time, within which spikes are characterized as binary values (1 represents the presence of a spike) \cite{rathi2020iclr}.

\begin{equation}
U_i(t)=U_i(t-1)+\sum_j W_{ij}{S_j(t)}-{V}\cdot{O_j(t-1)}
\label{eq:IF_discrete}
\end{equation}

\begin{equation}
O_i(t-1)=
\begin{cases}
    1, & \text{if } U_i(t-1)>V\\
    0, & \text{otherwise}
\end{cases}
\label{eq:IF_out_spike}
\end{equation}
\noindent
$O_i(t)$ is the output spike at time step $t$. Note that the third term in Eq. \ref{eq:IF_discrete} exhibits soft reset by reducing the membrane potential $U_i$ by the threshold $V$ at time step $t$, if an output spike is generated at the $(t-1)^{th}$ time step. Alternatively, hard reset implies resetting $U_i$ to $0$. Soft reset minimizes the information loss by allowing the spiking neuron to carry forward the surplus potential above the firing threshold to the next time step \cite{rathi2020iclr}.
\subsection{SNN Training Techniques}
Recent research on training supervised deep SNNs can be broadly divided into three categories: 1) ANN-to-SNN conversion-based training, 2) Spike timing dependent backpropagation (STDB), and 3) Hybrid training.
\subsubsection{ANN-to-SNN Conversion}\label{subsec:ann_snn_conv}

Recent works have demonstrated that SNNs can be efficiently converted from ANNs by approximating the activation value of ReLU neurons with the firing rate of spiking neurons \cite{dsnn_conversion1, dsnn_conversion_cont,dsnn_conversion_ijcnn,dsnn_conversion_abhronilfin,dsnn_conversion5}. This technique uses standard backpropagation-based training for the ANN models and helps an iso-architecture SNN achieve superior classification accuracy in image recognition tasks \cite{dsnn_conversion_cont,dsnn_conversion_abhronilfin}. 
However, the SNNs resulting from these ANN-SNN conversion algorithms require an order of magnitude higher latency compared to other training techniques \cite{dsnn_conversion_abhronilfin}. 
In this work, we use ANN-SNN conversion as an initial step in Q-STDB because it is of relatively low complexity and yields high classification accuracy on deep networks.

\subsubsection{STDB}
\label{subsec:snn_backprop}

The threshold comparator in the IF neuronal model yields a discontinuous and thus non-differentiable function, making it incompatible with the powerful gradient-descent based learning methods. Consequently, several approximate training methodologies have been proposed to overcome the challenges associated with non-differentiability \cite{lee_dsnn, panda2016_sup, bellec_2018long, neftci_surg}. The key idea of these works is to approximate the spiking neuron functionality with a continuous differentiable model or use surrogate gradients as an approximate version of the real gradients to perform gradient descent based training. Unfortunately, SNNs trained using this approach generally require a large number of time steps, in the order of few hundreds, to process an input. As a result, the backpropagation step requires the gradients of the unrolled SNN to be integrated over all these time steps. This multiple-iteration backpropagation-through-time (BPTT) coupled with the exploding memory complexity has hindered the applicability of surrogate gradient based learning methods to deep convolutional architectures. 

\subsubsection{Hybrid Training}
\label{subsec:snn_hybrid}
A recent paper \cite{rathi2020iclr} proposed a hybrid training methodology where the ANN-SNN conversion is performed as an initialization step and is followed by an approximate gradient descent algorithm. The authors observed that combining the two training techniques helps the SNNs converge within a few epochs while requiring fewer time steps.  Another recent paper \cite{rathi2020diet} proposed a training scheme for deep SNNs in which the membrane leak and the firing threshold along with other network parameters (weights) are updated at the end of every batch via gradient descent after ANN-SNN conversion. Moreover, \cite{rathi2020diet} applied direct-input encoding where the pixel intensities of an image are fed into the SNN input layer as fixed multi-bit values each timestep to reduce the number of required fewer time steps needed to achieve SOTA accuracy. Thus, the first convolutional layer composed of LIF neurons acts as both a feature extractor and spike-generator.
This is similar to rate-coding except that the spike-rate of the first hidden layer is a function of its weights, membrane leak, and threshold parameters that are all learned by gradient descent. 
This work extends these hybrid learning techniques by incorporating weight quantization, as defined below.

\section{Proposed Quantized SNN Training Method}\label{sec:snn_training}

In this section, we evaluate and compare the different choices for SNN quantization in terms of compute efficiency and model accuracy. We then incorporate the chosen quantization technique into STDB, which we refer to as Q-STDB.

\subsection{Study of Quantization Choice}\label{subsec:quantization_choice} 

Uniform quantization transforms a weight element $w\in [w_{min},w_{max}]$ to a range $[-2^{b-1},2^{b-1}-1]$ where $b$ is the bit-width of the quantized integer representation. There are primarily two choices for the above transformation, known as \textit{affine} and \textit{scale} quantization. Detailed descriptions of these two types of quantization can be found in \cite{jain2020trained}.

Our key motivation for SNN weight quantization is the hardware acceleration of inference using energy-efficient integer or fixed-point computational units implemented as crossbar array based processing-in-memory (PIM) accelerators. Note that the six transistor SRAM array based in-memory computing requires low-precision weights for multiply-and-accumulate (MAC) operations due to low density of the bit-cells. Previous research \cite{rathi2017stdp,sulaiman2020wquantize} have proposed post-training SNN quantization tailored towards unsupervised learning, which may not scale to 
complex vision tasks without requiring high-precision ($\geq{8}$ bits).
In contrast, in this work, we propose quantization-aware training, where the weights are fake quantized (see \cite{jain2020trained}) in the forward path computations, while the gradients and weight updates are calculated using the full precision weights. 
There are several choices for sharing quantization parameters among the tensor elements in a SNN. We refer to this choice as quantization granularity. We employ per-tensor (or per-layer) granularity where the same quantization parameters are shared by all elements in the tensor, because this reduces the computational cost compared to other granularity choices with no impact on model accuracy. Activations are similarly quantized, but only in the SNN input layer, since they are binary spikes in the remaining layers.

To evaluate the compute cost, let us consider a 3-D convolutional layer $l$, the dominant layer in HSI classification models, that performs a tensor operation $O_l=X_l\circledast W_l$ where $X_l\in \mathbb{R}^{H_l^i\times{W_l^i}\times{C_l^i}\times{D_l^i}}$ is the input activation tensor, $W_l\in \mathbb{R}^{k_l^x\times{k_l^y}\times{k_l^z}\times{C_l^i}\times{C_l^o}}$ and $O^l\in \mathbb{R}^{H_l^o\times{W_l^o}\times{C_l^o}\times{D_l^o}}$ is the output activation tensor, where $H_l^i$, $W_l^i$, $C_l^i$, $D_l^i$ are the input height, width, channel size, and spectral size, respectively. Similarly, $H_l^o$, $W_l^o$, $C_l^o$ and $D_l^o$ are the output height, width, number of filters and the output spectral size, respectively, and $k_l^x$, $k_l^y$, $k_l^z$ represent the filter size in the three spatial dimensions. The result of the real-valued operation $O_l = X_l\circledast W_l$ can be approximated with quantized tensors $X_l^Q$ and $W_l^Q$, by first dequantizing them producing $\hat{X_l}$ and $\hat{W_l}$ respectively, and then performing the convolution. Note that both $X_l^Q$ and $W_l^Q$ have similar dimensions as $X_l$ and $W_l$ respectively. Assuming the tensors are scale-quantized per layer,
\begin{equation}\label{eq:quant_tensor}
    O_l=X_l\circledast W_l\approx \hat{X_l}\circledast\hat{W_l}={X_l^Q\circledast W_l^Q}\cdot(\frac{1}{s_s^X\cdot{s_s^W}})
\end{equation}
\noexpand
where $s_s^X$ and $s_s^W$ are scalar values for scale quantization representing the levels of the input and weight tensor respectively. Hence, scale quantization results in an integer convolution, followed by a point-wise floating-point multiplication for each output element. Given that a typical convolution operation involves a few hundred MAC operations (accumulate for binary spike inputs) to compute an output element, a single floating-point operation for the scaling shown in Eq. \ref{eq:quant_tensor} is a negligible computational cost. Note that $X_l$ only needs to be quantized if $l$ is the input layer. In all other cases, $X_l^Q=X_l$ and $s_s^X=1$.

Although both affine and scale quantization enable the use of low-precision arithmetic, affine quantization results in higher computationally expensive inference as shown below. 
\begin{align}\label{eq:quant_tensor_affine}
    O_l&\approx\frac{X_l^Q-z_a^X}{s_a^X}\circledast \frac{W_l^Q-z_a^W}{s_a^W} \notag \\ &=\frac{({X_l^Q}\circledast{W_l^Q}-z_a^X\circledast(W_l^Q-z_a^W)-X_l^Q\circledast z_a^W)}{s_a^X\cdot{s_a^W}}
\end{align}
\noexpand
where $z_a^X$ and $z_a^W$ are tensors of sizes equal to that of $X_l^Q$ and $W_l^Q$ respectively that consist of repeated elements of the scalar zero-values of the input activation and weight tensor respectively. On the other hand, $s_a^X$ and $s_a^W$ are the corresponding scale values. 
The first term in the numerator of Eq. \ref{eq:quant_tensor_affine} is the integer convolution operation similar to the one performed in scale quantization shown in Eq. \ref{eq:quant_tensor}. The second term contains integer weights and zero-points, which can be computed offline, and adds an element-wise addition during inference. The third term, however, involves the quantized activation $X_l^Q$, which cannot be computed offline. This extra computation, depending on the implementation, can introduce considerable overhead, reducing or even eliminating the throughput and energy advantage that low precision PIM accelerators offer over floating-point MAC units. Hence, we use scale quantization during inference. 

Note, however, that our experiments detailed in Section \ref{sec:expt} show that using scale quantization during SNN training degrades the test accuracy significantly. Hence, we propose that training should use affine quantization of both the weights and input layer activations. Note that for a integer math unit or PIM accelerator, we do not necessarily need to quantize the SNN membrane potentials which are obtained as results of the accumulate operations of the weight elements. This is because the membrane potentials only need to be compared with the threshold voltage once for each time step, which consumes negligible energy, and can be performed using high precision fixed-point comparators (in the periphery of the crossbar array for PIM accelerators).  However, quantizing the potentials can reduce the data movement cost as discussed in Section \ref{sub:flops}.

\begin{figure}[t!]
\begin{center}
\includegraphics[width=0.46\textwidth]{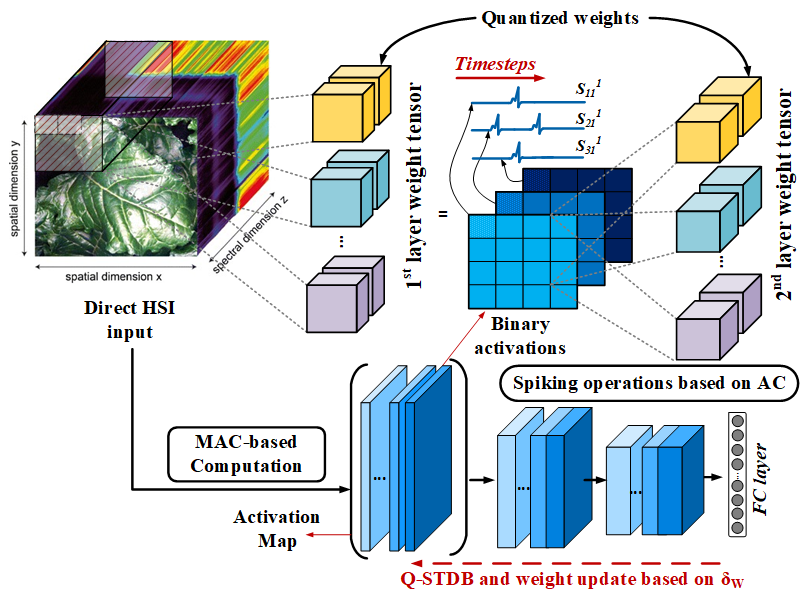}
\end{center}
\vspace{-0.2cm}
\caption{Proposed SNN training framework details with 3-D convolutions.}
\label{fig:proposed_training}
\vspace{-0.3cm}
\end{figure}

\subsection{Q-STDB based Training} 

In this subsection, we derive the expressions to compute the gradients of the parameters at all layers for our training framework. Our framework, which is illustrated in Fig. \ref{fig:proposed_training}, incorporates the quantization methodology described above into the STDB technique used to train SNNs \cite{rathi2020diet}, where the spatial and temporal credit assignment is performed by unrolling the network in time and employing BPTT.   \\
\textit{Output Layer:} The neuron model in the output layer $L$ only accumulates the incoming inputs without any leakage, does not generate an output spike, and is described by
\begin{align}
    \mbox{\boldmath $u$}_L^t &=\mbox{\boldmath $u$}_L^{t-1}+\hat{\mbox{\boldmath $w$}_L} \mbox{\boldmath $o$}_{L-1}^t \label{eq:lif_output} 
\end{align}
\noexpand
where $N$ is the number of output labels, $\mbox{\boldmath $u$}_L$ is a vector containing the membrane potential of $N$ output neurons, $\mbox{\boldmath $\hat{w}$}_L$ is the fake quantized weight matrix connecting the last two layers ($L$ and $L{-}1$), and $\mbox{\boldmath $o$}_{L-1}$ is a vector containing the spike signals from layer $(L{-}1)$. The loss function is defined on $\mbox{\boldmath $u$}_L$
at the last time step $T$. We employ the cross-entropy loss and compute the softmax of $\mbox{\boldmath $u$}_L^T$. The output of the network is passed through a softmax layer that outputs a probability distribution. The loss function $\mathcal{L}$ is defined as the cross-entropy between the true output ($y$) and the SNN's predicted distribution ($p$).

\begin{figure}[t!]
\begin{center}
\includegraphics[width=0.45\textwidth]{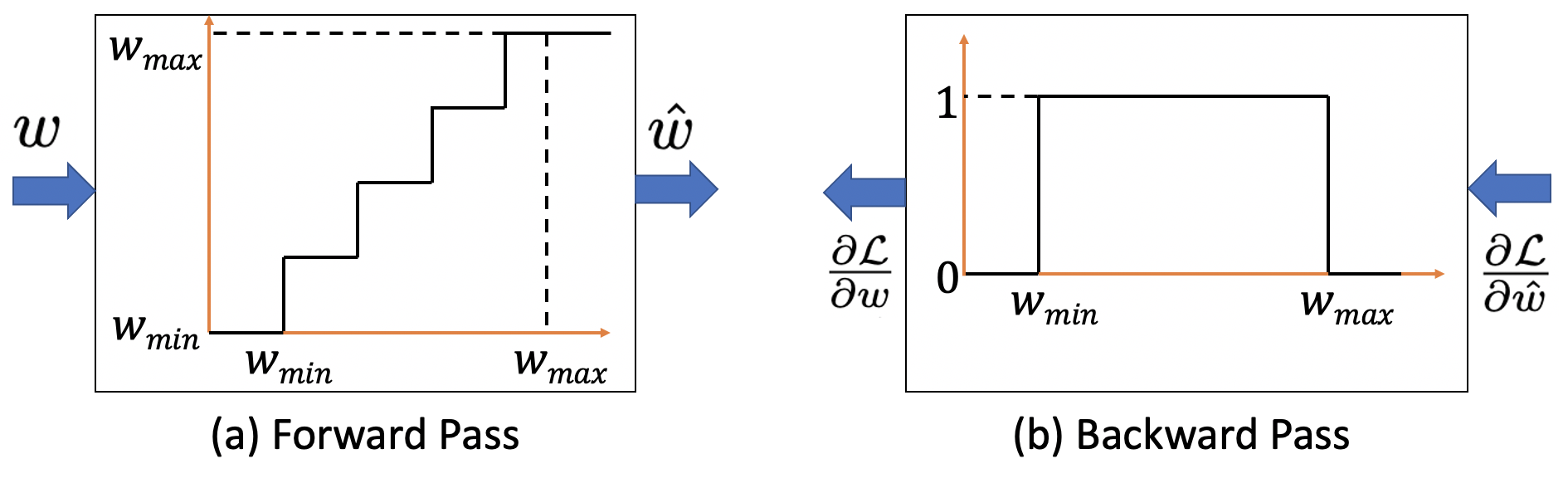}
\end{center}
\vspace{-0.6cm}
\caption{Fake quantization forward and backward pass with straight through estimator (STE) approximation}
\label{fig:weight_quantization}
\vspace{-0.6cm}
\end{figure}

\begin{equation}
    \mathcal{L}=-\sum_{i=1}^N{y_ilog({p_i})}, \quad  {p_i}=\frac{e^{u_i^T}}{\sum_{j=1}^{N}e^{u_j^T}},
\end{equation}
\noexpand
The derivative of the loss function with respect to the membrane potential of the neurons in the final layer is described by $\frac{\partial\mathcal{L}}{\partial\mbox{\boldmath $u$}_L^T}=(\mbox{\boldmath ${p}$}-\mbox{\boldmath $y$})$, where ${\mbox{\boldmath $p$}}$ and  ${\mbox{\boldmath $y$}}$ are vectors containing the softmax and one-hot encoded values of the true label respectively. To compute the gradient at the current time step, the membrane potential at the previous step is considered as an input quantity \cite{rathi2020diet}. With the weights being fake quantized, gradient descent updates the network parameters $\mbox{\boldmath $w$}_L$  of the output layer as

\begin{align}
\mbox{\boldmath $w$}_L&=\mbox{\boldmath $w$}_L-\eta\Delta{\mbox{\boldmath {$w$}}_L}\\
\Delta{\mbox{\boldmath $w$}_L}&=\sum_{t}\frac{\partial\mathcal{L}}{\partial \mbox{\boldmath $w$}_L}=\sum_{t}\frac{\partial\mathcal{L}}{\partial\mbox{\boldmath $u$}_L^t}\frac{\partial\mbox{\boldmath $u$}_L^t}{\partial \mbox{\boldmath $\hat{w}$}_L}\frac{\partial\mbox{\boldmath $\hat{w}$}_L}{\partial \mbox{\boldmath $w$}_L} \notag \\
&=\frac{\partial\mathcal{L}}{\partial\mbox{\boldmath $u$}_L^T}\sum_{t}\frac{\partial\mbox{\boldmath $u$}_L^t}{\partial \mbox{\boldmath $\hat{w}$}_L}\frac{\partial\mbox{\boldmath $\hat{w}$}_L}{\partial \mbox{\boldmath $w$}_L}\approx(\mbox{\boldmath ${p}$}-\mbox{\boldmath $y$})\sum_{t}\mbox{\boldmath $o$}_{L-1}^t \label{eq:wgradient}  \\
\frac{\partial\mathcal{L}}{\partial \mbox{\boldmath $o$}_{L-1}^t}&=\frac{\partial\mathcal{L}}{\partial \mbox{\boldmath $u$}_L^t}\frac{\partial \mbox{\boldmath $u$}_L^t}{\partial \mbox{\boldmath $o$}_{L-1}^t}=(\mbox{\boldmath ${p}$}-\mbox{\boldmath $y$})\mbox{\boldmath $\hat{w}$}_L
\end{align}
\noexpand
where $\eta$ is the learning rate (LR). Note that the derivative of the fake quantization function of the weights ($\frac{\partial\mbox{\boldmath $\hat{w}$}_L}{\partial \mbox{\boldmath $w$}_L}$) is undefined at the step boundaries and zero everywhere, as shown in Fig. \ref{fig:weight_quantization}(a). Our training framework addresses this challenge by using the Straight-through Estimator (STE) \cite{courbariaux2016binarized}, which approximates the derivative to be equal to 1 for inputs in the range $[w_{min}, w_{max}]$ as shown in Fig. \ref{fig:weight_quantization}(b), where $w_{min}$ and $w_{max}$ are the minimum and maximum values of the weights in a particular layer. Note that $w_{min}$ and $w_{max}$ are updated at the end of every mini-batch to ensure all the weights lie between $w_{min}$ and $w_{max}$ during the forward and backward computations in each training iteration. Hence, we use $\frac{\partial\mbox{\boldmath $\hat{w}$}_L}{\partial \mbox{\boldmath $w$}_L}\approx 1$ to compute the loss gradients in Eq. \ref{eq:wgradient}. \\
\textit{Hidden layers}: The neurons in the hidden convolutional and fully-connected layers are defined by the quantized LIF model as
\begin{align}
    \mbox{\boldmath $u$}_l^t&=\lambda_l{\mbox{\boldmath $u$}_l^{t-1}}+\mbox{\boldmath $\hat{w}$}_l{\mbox{\boldmath $o$}_{l-1}^t}-v_l\mbox{\boldmath $o$}_l^{t-1}\label{eq:lif_1} \\
    \mbox{\boldmath $z$}_l^t&=\frac{\mbox{\boldmath $u$}_l^t}{v_l}-1, \quad \mbox{\boldmath $o$}_l^t = \begin{cases}
    1, & \text{if } \mbox{\boldmath $z$}_l^t>0  \\
    0, & \text{otherwise }
       \end{cases} \label{eq:lif_2}
    \end{align}
    \noexpand
\noexpand  
where $\lambda_l$ and $v_l$ represent the leak and threshold potential for all neurons in layer $l$. All neurons in a layer possess the same leak and threshold value. This reduces the number of trainable parameters and we did not observe any significant improvement by assigning individual threshold/leak to each neuron. Given that the threshold is same for all neurons in a particular layer, it may seem redundant to train both the weights and threshold together. However, we observe that the number of time steps required to obtain the state-of-the-art classification accuracy decreases with this joint optimization. We hypothesize that this is because the optimizer is able to reach an improved local minimum when both parameters are tunable. The weight update in Q-STDB is calculated as
\begin{align}
    \Delta{w_l}&=\sum_{t}\frac{\partial\mathcal{L}}{\partial w_l}=\sum_{t}\frac{\partial\mathcal{L}}{\partial\mbox{\boldmath $z$}_l^t}\frac{\partial\mbox{\boldmath $z$}_l^t}{\partial\mbox{\boldmath $o$}_l^t}\frac{\partial\mbox{\boldmath $o$}_l^t}{\partial\mbox{\boldmath $u$}_l^t}\frac{\partial\mbox{\boldmath $u$}_l^t}{\partial \mbox{\boldmath $\hat{w}$}_l}\frac{\partial\mbox{\boldmath $\hat{w}$}_l}{\partial\mbox{\boldmath $w$}_l} \notag \\
    &\approx\sum_{t}\frac{\partial\mathcal{L}}{\partial\mbox{\boldmath $z$}_l^t}\frac{\partial\mbox{\boldmath $z$}_l^t}{\partial\mbox{\boldmath $o$}_l^t}\frac{1}{v_l}\mbox{\boldmath $o$}_{l-1}^t\cdot{1}
\end{align}
\noexpand
where $\frac{\partial\mbox{\boldmath $\hat{w}$}_l}{\partial\mbox{\boldmath $w$}_l}$ and $\frac{\partial\mbox{\boldmath $z$}_l^t}{\partial\mbox{\boldmath $o$}_l^t}$ are the two discontinuous gradients. We calculate the former using STE described above, while the latter is approximated using surrogate gradient  \cite{bellec_2018long} shown below.
\begin{align}
    \frac{\partial\mbox{\boldmath $z$}_l^t}{\partial\mbox{\boldmath $o$}_l^t}=\gamma\cdot{max(0,1-|\mbox{\boldmath $z$}_l^t|)}
\end{align}
\noexpand
Note that $\gamma$ is a hyperparameter denoting the maximum value of the gradient. The threshold and leak update is computed similarly using BPTT \cite{rathi2020diet}.
\noexpand

\section{Proposed Architectures}\label{sec:arch}

We developed two models, a 3-D and a hybrid fusion of 3-D and 2-D convolutional architectures, that are inspired by the recently proposed CNN models \cite{hamida2018-3d,luo2018hsi, roy2020hybrid} used for HSI classification and compatible with our ANN-SNN conversion framework. We refer to the two models CNN-3D and CNN-32H.
The models are trained without the bias term because it complicates parameter space exploration which increases the conversion difficulty and tends to increase conversion loss. The absence of the bias term implies that batch normalization \cite{bjorck2018understanding} cannot be used as a regularizer during the training process. Instead, we use dropout \cite{srivastava_2014} as the regularizer for both ANN and SNN training. Moreover, our models employ ReLU nonlinearity after each convolutional and linear layer (except the classifier layer) to further decrease the conversion loss due to the similarity between ReLU and LIF neurons. Also, our pooling operations use average pooling because for binary spike based activation layers, max pooling incurs significant information loss. Additionally, we modified the number of channels and convolutional layers to obtain a reasonable tradeoff between accuracy and compute efficiency. 2-D patches of sizes $5{\times}5$ and $3{\times}3$ were extracted for CNN-3D and CNN-32H respectively, without any reduction in dimensionality from each dataset. Higher sized patches increase the computational complexity without any significant improvement in test accuracy. Our model architectures are explicitly described in Table \ref{tab:models_cnn1and2}.

\vspace{-0.2cm}

\section{Experiments}\label{sec:expt}

\subsection{Datasets}

We used three publicly available datasets, namely Indian Pines, Pavia University, and Salinas scene. A brief description follows for each one \cite{hyperspectraldatasets}.

\textit{Indian Pines}: The Indian Pines (IP) dataset consists of $145{\times}145$ spatial pixels and $220$ spectral bands in a range of $400-2500$ nm. It was captured using the AVIRIS sensor over North-Western Indiana, USA, with a ground sample distance (GSD) of $20$ m and has $16$ vegetation classes. \\
\textit{Pavia University}: The Pavia University (PU) dataset consists of hyperspectral images with $610{\times}340$ pixels in the spatial dimension, and $103$ spectral bands, ranging from $430$ to $860$ nm in wavelength. It was captured with the ROSIS sensor with GSD of $1.3$ m over the University of Pavia, Italy. It has a total of 9 urban land-cover classes. \\
\textit{Salinas Scene}: The Salinas Scene (SA) dataset contains images with $512{\times}217$ spatial dimension and $224$ spectral bands in the wavelength range of $360$ to $2500$ nm. The $20$ water absorbing spectral bands have been discarded. It was captured with the AVIRIS sensor over Salinas Valley, California with a GSD of $3.7$ m. In total $16$ classes are present in this dataset. 

For preprocessing, images in all the data sets are normalized to have a zero mean and unit variance. 
For our experiments, all the samples are randomly divided into two disjoint training and test sets. The limited 40\% samples are used for training and the remaining 60\% for performance evaluation.

\begin{table}
\caption{Model architectures employed for CNN-1 and CNN-2. Every convolutional and linear layer is followed by ReLU non-linearity. The last classifier layer is not shown.}
\begin{center}
\scriptsize\addtolength{\tabcolsep}{-1.5pt}
\begin{tabular}{|c|c|c|c|c|c|}
\hline
Layer & Number of & Size of & Stride & Padding & Dropout   \\
type & filters & each filter & value & value & value \\
\hline
\hline
 \multicolumn{6}{|c|}{Architecture : CNN-3D} \\
\hline
\hline
3-D Convolution & 20  & (3,3,3) & (1,1,1) & (0,0,0) & -  \\
\hline
3-D Convolution & 40  & (3,1,1) & (2,1,1) & (1,0,0) & -  \\
\hline
3-D Convolution & 84 & (3,3,3) & (1,1,1) & (1,0,0) & -  \\
\hline
3-D Convolution & 84 & (3,1,1) & (2,1,1) & (1,0,0) & - \\
\hline
3-D Convolution & 84 & (3,1,1) & (1,1,1) & (1,0,0) & - \\
\hline
3-D Convolution & 84 & (2,1,1) & (2,1,1) & (1,0,0) & - \\
\hline
\hline
 \multicolumn{6}{|c|}{Architecture : CNN-32H} \\
\hline
\hline
3-D Convolution & 90  & (18,3,3) & (7,1,1) & (0,0,0) & -  \\
\hline
2-D Convolution & 64  & (3,3) & (1,1) & (0,0) & -  \\
\hline
2-D Convolution & 128  & (3,3) & (1,1) & (0,0) & -  \\
\hline
Avg. Pooling & - & (4,4) & (4,4) & (0,0) & - \\
\hline
Dropout & - & - & - & - & 0.2 \\
\hline
Linear & 30998528 & - & - & - & - \\
\hline
\end{tabular}
\end{center}
\label{tab:models_cnn1and2}
\vspace{-3mm}
\end{table}

\subsection{Experimental Setup}
\label{subsec:setup}


\subsubsection{ANN Training}
\label{subsubsec:model_and_ann}
We performed full-precision ANN training for $100$ epochs using the standard SGD optimizer with an initial learning rate (LR) of $0.01$ that decayed by a factor of 0.1 after $60$, $80$, and $90$ epochs. 

\subsubsection{Conversion and SNN Training}

We first examine the distribution of the neuron input values over the total number of time steps across all neurons of the first layer for a small batch of HSI images (of size $50$ in our case) and set the layer threshold to the $99.7$ percentile of the scaled value of the evaluated threshold \cite{rathi2020diet}. In our experiments we scale the initial thresholds by 0.8. Similarly, we then compute the thresholds of the subsequent layers sequentially by examining the distribution of their input 
values. Note that we use $100$ time steps to evaluate the thresholds, while the SNN training and inference are performed with only $5$ time steps. We keep the leak of each layer set to unity while evaluating initial thresholds. At the start of SNN training, we initialize the weights with those from the trained ANN and initialize the leak parameters to $1.0$. 
We then perform the quantization-aware SNN training described in Section \ref{sec:snn_training} for another $100$ epochs. We set $\gamma$ = $0.3$ \cite{bellec_2018long} and used the ADAM optimizer with a starting LR of $10^{-4}$ which decays by a factor of $0.5$ after $60$, $80$, and $90$ epochs. 

\begin{figure}[!t]
\includegraphics[width=0.48\textwidth]{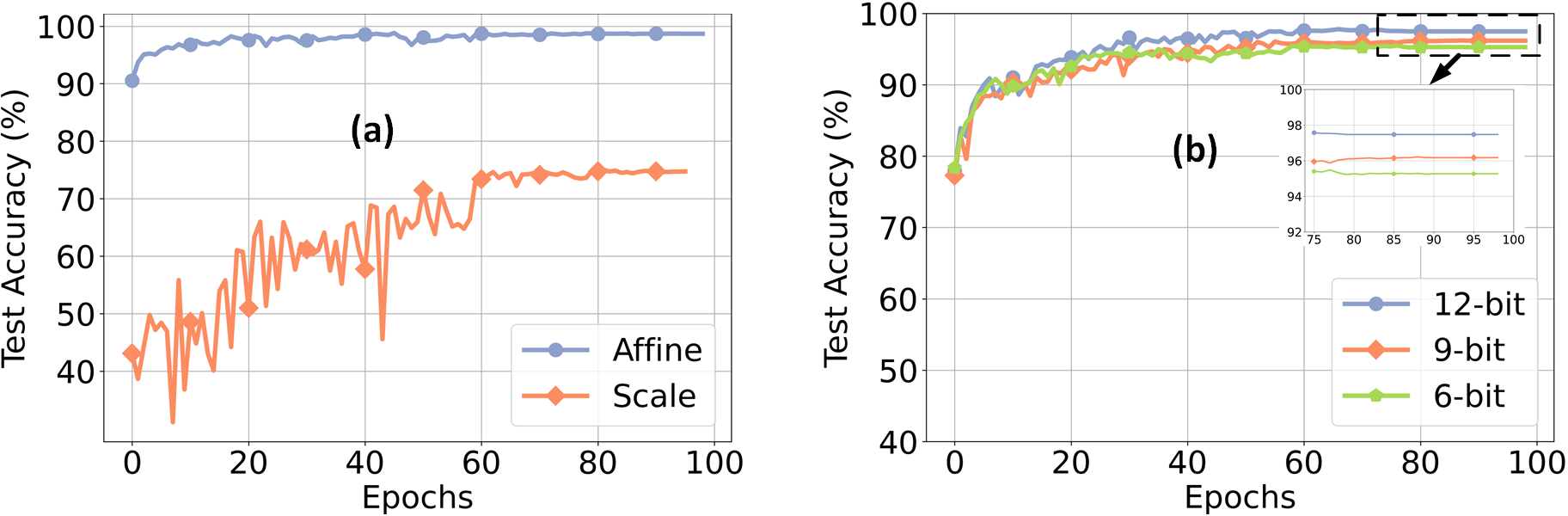}
\centering
\caption{(a) Test accuracies for affine and scale quantization with CNN-3D over IP dataset (b) Test accuracies with 6, 9 and 12-bit weight precisions for post-training quantization with CNN-32H on IP dataset.}
\label{fig:vgg_diff_bit_quant_acc_plots}
\vspace{-2mm}
\end{figure}

\begin{table*}[!t]
\caption{Model performances with Q-STDB based training on IP, PU, and SS datasets for CNN-3D and CNN-32H after a) ANN training, b) ANN-to-SNN conversion, c) 32-bit SNN training, d) 4-bit SNN training, e) 5-bit SNN training, and f) 6-bit SNN training.}
\begin{center}
\scriptsize\addtolength{\tabcolsep}{-3.5pt}
\begin{tabular}{|c|c|c|c|c|c|c|c|c|c|c|c|c|c|c|c|c|c|c|}
\hline
  & \multicolumn{3}{|c|}{A. ANN} & \multicolumn{3}{|c|}{B. Accuracy after} & \multicolumn{3}{|c|}{C. Accuracy after}  & \multicolumn{3}{|c|}{D. Accuracy after} & \multicolumn{3}{|c|}{E. Accuracy after} & \multicolumn{3}{|c|}{F. Accuracy after}  \\
 Dataset & \multicolumn{3}{|c|}{accuracy ($\%$)} & \multicolumn{3}{|c|}{ANN-to-SNN conv. ($\%$)} &  \multicolumn{3}{|c|}{FP SNN training ($\%$)}    & \multicolumn{3}{|c|}{\textbf{4}-bit SNN training (\%)} & \multicolumn{3}{|c|}{\textbf{5}-bit SNN training (\%)} & \multicolumn{3}{|c|}{\textbf{6}-bit SNN training (\%)} \\
\cline{2-19}
 &  OA & AA & Kappa &  OA & AA & Kappa  & OA & AA & Kappa &  OA & AA & Kappa & OA & AA & Kappa & OA & AA & Kappa \\ 
\hline
\hline
 \multicolumn{19}{|c|}{Architecture : CNN-3D} \\
\hline
\hline
 IP  & 98.86 & 98.42 & 98.55 & 57.68 & 50.88 & 52.88 &  98.92 & 98.76 & 98.80 & 97.08 & 95.64 & 95.56 & 98.38 & 97.78 & 98.03 & 98.68 & 98.34 & 98.20 \\ 
\hline
PU  & 99.69  & 99.42 & 99.58 & 91.16 & 88.84 & 89.03 & 99.47 & 99.06 & 99.30 & 98.21 & 97.54 & 97.75  & 99.26 & 98.48 & 98.77 & 99.50 & 99.18 & 99.33  \\
\hline
SS  & 98.89  & 98.47 & 98.70 & 81.44 & 76.72 & 80.07 & 98.49 & 97.84  & 98.06 & 96.47 & 93.16 & 94.58 & 97.25 & 95.03 & 95.58 & 97.95 & 97.09 & 97.43 \\
\hline
 \multicolumn{19}{|c|}{Architecture : CNN-32H} \\
\hline
\hline
IP  & 97.60 & 97.08 & 97.44 & 70.88 & 66.56 & 67.89 & 97.27 & 96.29 & 96.35 & 96.63 & 95.81 & 95.89 & 97.23 & 96.08 & 96.56 & 97.45 & 96.73 & 96.89 \\ 
\hline
PU  & 99.50  &  99.09 & 99.30 & 94.96 & 90.12 & 93.82 & 99.38  & 98.83 & 99.13 & 99.17 & 98.41 & 98.68 & 99.25 & 98.84 & 98.86 & 99.35 & 98.88 & 98.95 \\
\hline
SS  & 98.88 & 98.39 & 98.67 & 88.16 & 84.19 & 85.28 & 97.92 & 97.20 & 97.34 & 97.34 & 96.32 & 96.77 & 97.65 & 96.81 & 96.97 & 97.99  & 97.26  & 97.38 \\
\hline
\end{tabular}
\end{center}
\label{tab:ann_snn_accuracy}
\vspace{-3mm}
\end{table*}

\begin{table}
\caption{Performance comparison of the SNNs trained with Q-STDB with state-of-the-art deep ANNs on IP and PU datasets}
\begin{center}
\scriptsize\addtolength{\tabcolsep}{-2.9pt}
\begin{tabular}{|c|c|c|c|c|c|}
\hline
Authors &  ANN/SNN & Architecture & OA ($\%$) & AA ($\%$) & Kappa ($\%$)  \\
\hline
\hline
 \multicolumn{6}{|c|}{Dataset : Indian Pines} \\
\hline
\hline
Alipour-Fard  & ANN & MSKNet & 81.73 & 71.4 & 79.2 \\
et al. (2020) \cite{fard2020kernel} &  &  &  & & \\
\hline
Song et al.  & ANN & DFFN & 98.52 & 97.69 & 98.32 \\
(2018)\cite{song2018feature} &  &  &  & & \\
\hline
Zhong et al. & ANN & SSRN & \textbf{99.19} & \textbf{98.93} & \textbf{99.07} \\
(2018) \cite{zhong2018hsi} &  &  &  & & \\
\hline
Roy et al. & ANN & HybridSN & 98.39 & 98.01 & 98.16 \\
(2020) \cite{roy2020hybrid} &  & &   &  &  \\
\hline
Hamida et al. & ANN & 6-layer  & 98.29 & 97.52 & 97.72 \\
(2018) \cite{hamida2018-3d}& SNN & 3D CNN & 95.88 & 94.26 & 95.34 \\
\hline
Luo et al. & ANN & Hybrid  & 96.15 & 94.96 & 95.73 \\
(2018) \cite{luo2018hsi}& SNN & CNN & 94.90  & 94.08 & 94.78 \\
\hline
\rowcolor{Gray}
This work & ANN & CNN-3D & 98.86 & 98.42 & 98.55 \\
\rowcolor{Gray}
 & SNN & & 98.68 & 98.34 & 98.20  \\
\rowcolor{Gray}
\hline
This work & ANN & CNN-32H & 97.60 & 97.08 & 97.44 \\
\rowcolor{Gray}
 & SNN & & 97.45 & 96.73 & 96.89  \\
\hline
\hline
 \multicolumn{6}{|c|}{Dataset : Pavia University} \\
\hline
\hline
Alipour-Fard  & ANN & MSKNet & 90.66 & 88.09 & 87.64 \\
et al. (2020) \cite{fard2020kernel} &  &  &  & & \\
\hline
Song et al.  & ANN & DFFN & 98.73 & 97.24 & 98.31 \\
(2018)\cite{song2018feature} &  &  &  & & \\
\hline
Zhong et al. & ANN & SSRN & 99.61 & \textbf{99.56} & 99.33 \\
(2018) \cite{zhong2018hsi} &  &  &  & & \\
\hline
Hamida et al. & ANN & 6-layer & 99.32 & 99.02 & 99.09 \\
(2018) \cite{hamida2018-3d}& SNN & 3D CNN & 98.55 & 98.02 & 98.28 \\
\hline
Luo et al. & ANN & Hybrid & 99.05 & 98.35 & 98.80 \\
(2018) \cite{luo2018hsi}& SNN & CNN & 98.40  & 97.66 & 98.21 \\
\hline
\rowcolor{Gray}
This work & ANN & CNN-3D & 99.69 & 99.42 & 99.58 \\
\rowcolor{Gray}
 & SNN & & 99.50 & 99.18 & 99.33  \\
 \hline
\rowcolor{Gray}
This work & ANN & CNN-32H & 99.50 & 99.09 & 99.30 \\
\rowcolor{Gray}
 & SNN & & 99.35 & 98.88 & 98.95  \\
 \hline
\end{tabular}
\end{center}
\label{tab:snn_comparison}
\vspace{-6mm}
\end{table}

\subsection{ANN \& SNN Inference Results}\label{subsec:inference_results}
\label{subsec:compression_res}

\begin{figure*}[t!]
\begin{center}
\includegraphics[width=1\textwidth]{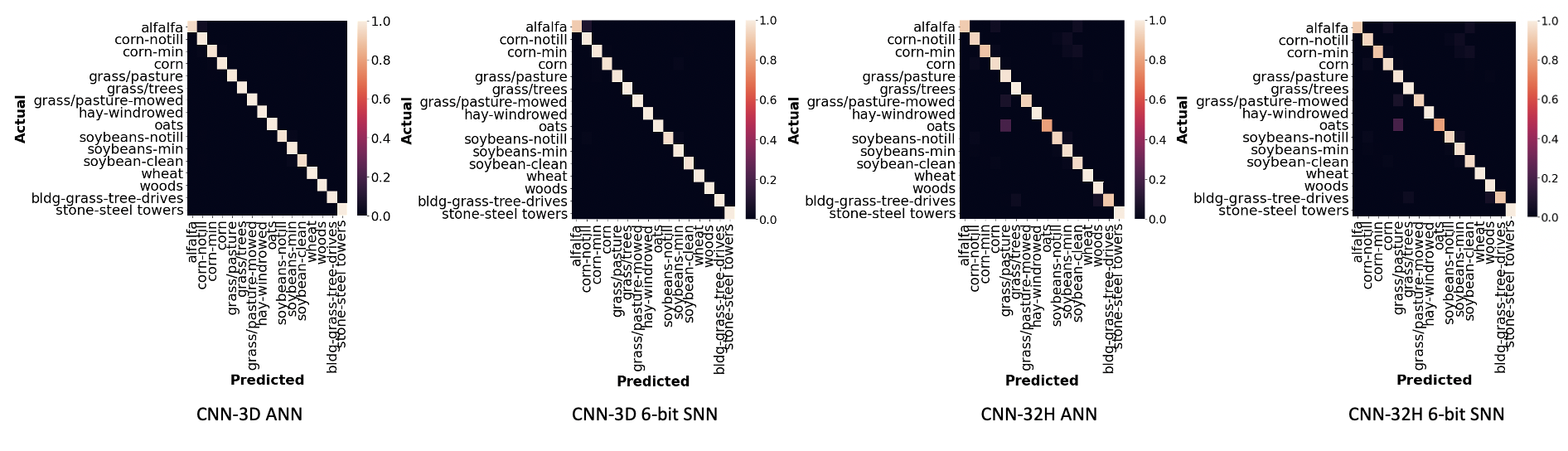}
\end{center}
\vspace{-0.6cm}
\caption{Confusion Matrix for HSI test performance of ANN and proposed 6-bit SNN over IP dataset for both CNN-3D and CNN-32H. The ANN and SNN confusion matrices look similar for both the network architectures. CNN-32H incurs a little drop in accuracy compared to CNN-3D due to shallow architecture.}
\label{fig:confusion_matrix}
\vspace{-0.6cm}
\end{figure*}

We have used the Overall Accuracy (OA), Average Accuracy (AA), and Kappa Coefficient evaluation measures to evaluate the HSI classification performance for our proposed architectures, similar to \cite{hamida2018-3d}. Here, OA represents the number of correctly classified samples out of the total test samples. AA represents the average of class-wise classification accuracies, and Kappa is a statistical metric used to assess the mutual agreement between the ground truth map and classification map. Column-$2$ in Table \ref{tab:ann_snn_accuracy} shows the ANN accuracies, column-$3$ shows the accuracy after ANN-SNN conversion with $50$ timesteps. Column-$4$ shows the accuracy when we perform our proposed training without quantization, while columns 5 to 7 shows the SNN test accuracies obtained with Q-STDB for different bit precisions (4 to 6 bits) of the weights. We observe that for all the datasets, SNNs trained with 6-bit weights result in $5.33\times$ reduction in bit-precision compared to full-precision (32-bit) models and perform almost at par with the full precision ANNs on both the architectures. 4-bit weights do not incur significant accuracy drop as well, and can be used for applications demanding high energy-efficiency and low latency.  
Fig. \ref{fig:confusion_matrix} shows the confusion matrix for the HSI classification performance of the ANN and proposed SNN over the IP dataset for both the architectures. Although the membrane potentials do not need to be quantized as described in Section \ref{sec:snn_training}, we observed that the model accuracy does not drop significantly even if we quantize them, and hence, the SNN results shown in Table \ref{tab:ann_snn_accuracy} correspond to 6-bit membrane potentials. Moreover, quantized membrane potentials can reduce the data movement cost as discussed in Section \ref{sub:flops}.

The performance of our ANNs and SNNs trained via Q-STDB are compared with the current state-of-the-art ANNs used for HSI classification in Table \ref{tab:snn_comparison}. Note that mere porting the ANN architectures used in \cite{hamida2018-3d,luo2018hsi} to SNNs, and performing  6-bit Q-STDB results in significant accuracy drop, and hence, shows the efficacy of our proposed architectures.

\subsubsection{Q-STDB vs Post-Training Quantization (PTQ)}

PTQ cannot always yield ultra low-precision SNNs with SOTA test accuracy. For example, for the IP dataset and CNN-32H architecture, the lowest bit precision of the weights that the SNNs can be trained with PTQ for no more than $1\%$ reduction in SOTA test accuracy is $12$, if we limit the total number of time steps to $5$. Fig. \ref{fig:vgg_diff_bit_quant_acc_plots}(b) shows the test accuracies for different bit precisions (${\leq}{12}$) of weights with PTQ on IP dataset. 
The weights can be further quantized to $8$-bits if we increase the time steps to $10$, which increases the latency. On the other hand, Q-STDB results in accurate (${\leq}{1}\%$ deviation from ANN accuracy) $6$ bit SNNs with only $5$ time steps, which improves both the energy-efficiency and latency. The energy-efficiency of our proposed architectures trained with Q-STDB are quantified in Section \ref{sec:edp_analysis}.  
\subsubsection{Affine vs Scale Quantization during Training}

As illustrated in Section \ref{sec:snn_training}, performing scale quantization during the forward path in training degrades the SNN accuracy significantly. Fig. \ref{fig:vgg_diff_bit_quant_acc_plots}(a) shows the test accuracies for affine and scale quantization during training with CNN-3D architecture on IP dataset.

\section{Improvement in Energy-Delay Product}\label{sec:edp_analysis}

\subsection{Spiking Activity}
\label{subsub:spiking_act} 
To model energy consumption, we assume a generated SNN spike consumes a fixed amount of energy \cite{dsnn_conversion1}. Based on this assumption, earlier works \cite{rathi2020iclr, dsnn_conversion_abhronilfin} have adopted the average spiking activity (also known as average spike count) of an SNN layer $l$, denoted ${\zeta}^l$, as a measure of compute-energy of the model. In particular, ${\zeta}^l$ is computed as the ratio of the total spike count in $T$ steps over all the neurons of the layer $l$ to the total number of neurons in that layer. Thus lower the spiking activity, the better the energy efficiency.

\begin{figure}[!t]
\includegraphics[width=0.32\textwidth]{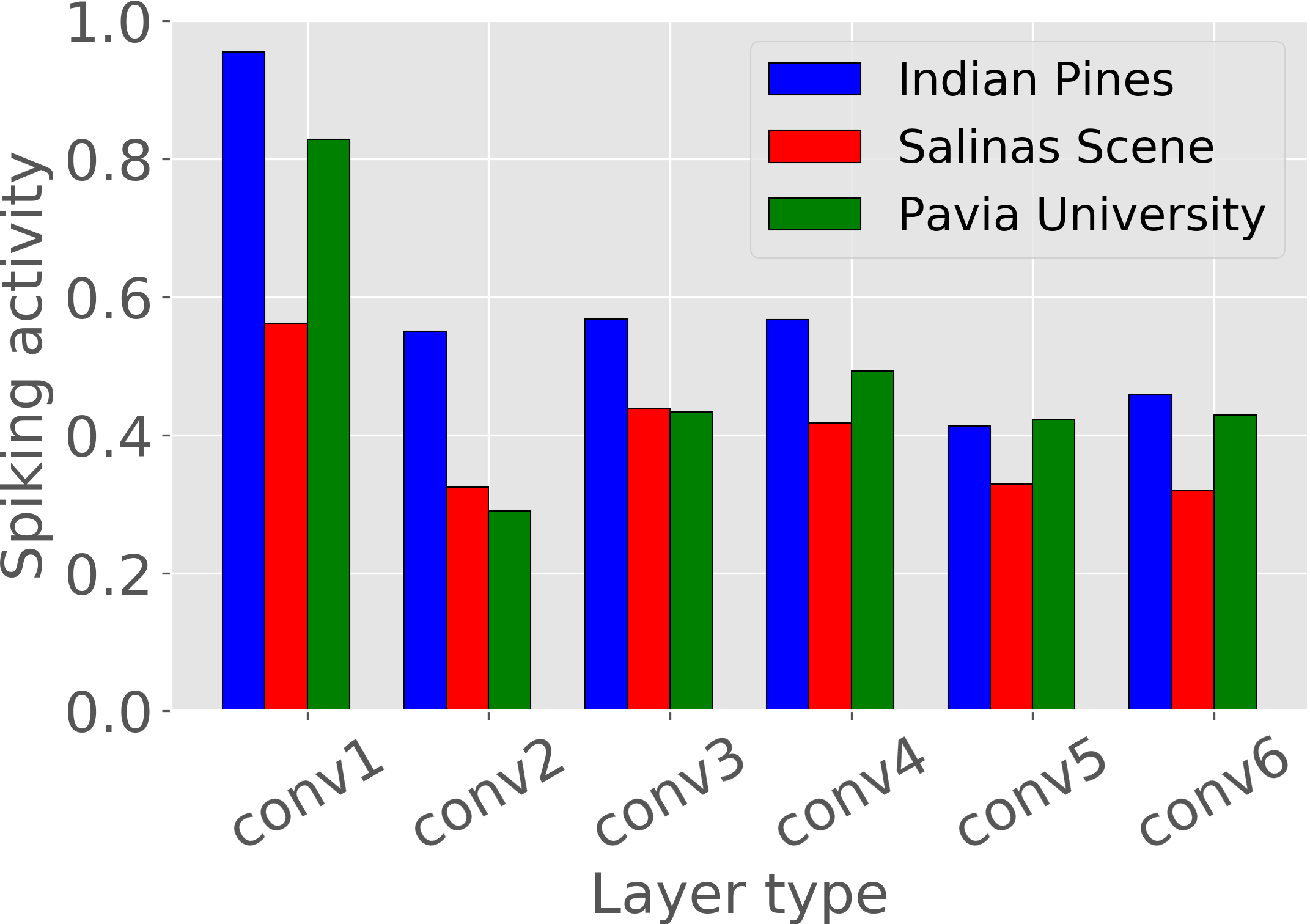}
\centering
\caption{Layerwise spiking activity plots for CNN-3D on Indian Pines, Salinas Scene and Pavia University datasets.}
\label{fig:vgg_sa_datasets}
\vspace{-4mm}
\end{figure}

Fig. \ref{fig:vgg_sa_datasets} shows the average number of spikes for each layer with Q-STDB when evaluated for $200$ samples from the IP testset for the CNN-3D architecture. Let the average be denoted by $\zeta_l$ which is computed by summing all the spikes in a layer over $5$ time steps and dividing by the number of neurons in that layer. For example, the average spike count of the $3^{rd}$ convolutional layer of the SNN is $0.568$, which implies that over a $5$ timestep period each neuron in that layer spikes $0.568$ times on average over all input samples. 

\subsection{Floating point operations count (FLOPs) \& Total Energy}\label{sub:flops}

Let us assume a 3-D convolutional layer $l$ having weight tensor $\textbf{W}^l \in \mathbb{R}^{k_l^x \times k_l^y \times k_l^z \times C_{l}^i \times C_{l}^o}$ that operates on an input activation tensor $\textbf{I}^l \in \mathbb{R}^{H_{l}^i \times W_{l}^i \times C_{l}^i \times D_{l}^i}$, where the notations are similar to the one used in Section \ref{sec:snn_training}. We now quantify the energy consumed to produce the corresponding output activation tensor $\textbf{O}^l \in \mathbb{R}^{H_{l}^o \times W_{l}^o \times C_{l}^o \times D_{l}^o}$ for an ANN and SNN, respectively. Our model can be extended to fully-connected layers with $f_{l}^{i}$ and $f_{l}^{o}$ as the number of input and output features respectively, and to 2-D convolutional layers, by shrinking a dimension of the feature maps.
In particular, for an ANN, the total number of FLOPS for layer $l$, denoted $F_{l}^{ANN}$, is shown in row $1$ of Table \ref{tab:flops} \cite{kundu2020pre, kundu2019psconv}. The formula can be easily adjusted for an SNN in which the number of FLOPs at layer $l$ is a function of the average spiking activity at the layer $(\zeta_l)$ denoted as $F_{l}^{SNN}$ in Table \ref{tab:flops}. Thus, as the activation output gets sparser, the compute energy decreases. 

For ANNs, FLOPs primary consist of multiply accumulate (MAC) operations of the convolutional and linear layers. On the contrary, for SNNs, except the first and last layer, the FLOPs are limited to accumulates (ACs) as the spikes are binary and thus simply indicate which weights need to be accumulated at the post-synaptic neurons. For the first layer, we need to use MAC units as we consume analog 
input\footnote{Note that for the hybrid coded data input we need to perform MAC at the first layer at $t=1$, and AC operation during remaining timesteps at that layer. For the direct coded input, only MAC during the $1^{st}$ timestep is sufficient, as neither the inputs nor the weights change during remaining timesteps (i.e. $ 5 \geq t \geq 2 $).} (at timestep one). Hence, the compute energy for an ANN $(E^{ANN})$ and an iso-architecture SNN model $(E^{SNN})$ can be written as 
\begin{align}
     E^{ANN} & =(\sum_{l=1}^{L}F^{SNN}_{l}){E_{MAC}}\\
   E^{SNN} &=(F^{ANN}_{1}){E_{MAC}}+ (\sum_{l=2}^{L}F^{SNN}_{l}){E_{AC}}
\end{align}
\noindent
where $L$ is the total number of layers. Note that $E_{MAC}$ and $E_{AC}$ are the energy consumption for a MAC and AC operation respectively. As shown in Table \ref{tab:fp_energy}, $E_{AC}$ is $\mathord{\sim}32 \times$ lower than $E_{MAC}$ \cite{horowitz20141} in $45$ nm CMOS technology for 32-bit precision. To compute $E_{MAC}$ and $E_{AC}$ for any arbitrary bit precision $Q$ (6-bits in our work), we use $E_{MAC}{\propto}Q^{1.25}$ \cite{moons2017qmac}, and $E_{AC}{\propto}Q$ \cite{simon2019qac}. These numbers may vary for different technologies, but generally, in most technologies, an AC operation is significantly less expensive than a MAC operation and its' energy scales close to linearly with bit precision. 

\begin{table}[!ht]
\caption{Floating point operations count (FLOPs) in Convolutional and Fully-connected layer for ANN and SNN models}
\scriptsize\addtolength{\tabcolsep}{-6pt}
\begin{center}
\begin{tabular}{|c|c|c|c|c|}
\hline
{Model} & \multicolumn{4}{|c|}{Number of FLOPs} \\ 
\cline{2-5}
        & Notation & 3-D Conv. layer $l$ & 2-D Conv. layer $l$   & FC layer $l$ \\
\hline
{$ANN$} &  $F_{l}^ANN$ & $k_l^x\times k_l^y\times k_l^z\times H_l^o\times$  & $(k_l)^2\times H_l^o\times W_l^o\times$ & $f_l^i\times f_l^o$ \\ 
 & & $W_l^o\times D_l^o\times C_l^o\times C_l^i$ &   $C_l^o\times C_l^i$ & \\   
\hline
{$SNN$} &  $F_{l}^SNN$ & $k_l^x\times k_l^y\times k_l^z\times H_l^o\times$ & $(k_l)^2\times H_l^o\times W_l^o\times $ & $f_l^i\times f_l^o\times \zeta_l$\\ 
  & &   $W_l^o\times D_l^o\times C_l^o\times C_l^i\times \zeta_l$    & $C_l^o\times C_l^i\times \zeta_l$ &  \\ 
\hline
\end{tabular}
\label{tab:flops}
\vspace{-3mm}
\end{center}
\end{table}

\begin{figure*}[!t]
\includegraphics[width=0.98\textwidth]{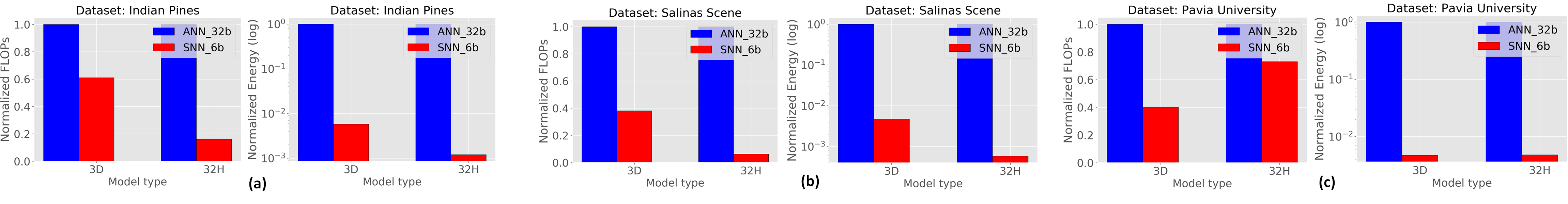}
\vspace{-2mm}
\centering
\caption{Comparison of FLOPs and compute energy of CNN-3D and CNN-32H between ANN and SNN models while classifying on (a) Indian Pines, (b) Salinas Scene and (c) Pavia University datasets, respectively.}
\label{fig:FLOPs_and_energy_plot}
\vspace{-4mm}
\end{figure*}

Fig. \ref{fig:FLOPs_and_energy_plot} illustrates the compute energy consumption and FLOPs for full precision ANN and 6-bit quantized SNN models of the two proposed architectures while classifying the IP, PU, and SS datasets, where the energy is normalized to that of an equivalent ANN. We also consider $6$-bit ANN models to compare the energy-efficiency of low-precision ANNs and SNNs. As seen in Fig. \ref{fig:FLOPs_and_energy_plot}, 6-bit ANN models are $12.5\times$ energy-efficient compared to 32-bit ANN models due to the similar factor of improvement in MAC energy (see Table \ref{tab:fp_energy}). Note that we can achieve the HSI test accuracies shown in Table \ref{tab:ann_snn_accuracy} with quantized ANNs as well.

The FLOPs for SNNs obtained by our proposed training framework is smaller than that for an ANN with similar number of parameters due to low spiking activity. Moreover, because the ACs consume significantly less energy than MACs for all bit precisions, SNNs are significantly more compute efficient. In particular, for CNN-3D on IP, our proposed SNN consumes $\mathord{\sim}199.3\times$ and $\mathord{\sim}15.9\times$ less compute energy than an iso-architecture full-precision and 6-bit ANN with similar parameters respectively. The improvements become ${\sim}560.6\times$ and ${\sim}44.8\times$ respectively on averaging across the two network architectures and three datasets. Note that we did not consider the memory access energy in our evaluation because it is dependent on the underlying system architecture. In general, SNNs incur significant data movement because both the membrane potentials and weights need to be fetched from the on-chip memory. Q-STDB addresses the memory cost by reducing their bit precisions by $5.33\times$ (see Section \ref{subsec:inference_results}) compared to full-precision models. Moreover, there have been many proposals to reduce the memory cost by data buffering \cite{shen_2017}, computing in non-volatile crossbar memory arrays \cite{chen2015crossbar}, and data reuse with energy-efficient dataflows \cite{eyeriss_v1}. All these techniques can be complemented with Q-STDB to further decrease the memory cost.

\begin{table}
\caption{Estimated energy costs for $32$ and $6$-bit MAC and AC operations in 45 $nm$ CMOS process}
\scriptsize\addtolength{\tabcolsep}{-4pt}
\begin{center}
\begin{tabular}{|c|c|c|}
\hline
{Bit-precision} & {Operation} & {Energy ($pJ$)} \\
\hline
32 & Mutiply-and-Accumulate (MAC) & $3.2$ \\
 & Accumulate (AC) & $0.1$ \\
\hline
6 & Mutiply-and-Accumulate (MAC) &  $0.26$ \\
 & Accumulate (AC)  & $0.02$ \\
\hline
\end{tabular}
\label{tab:fp_energy}
\vspace{-0.6cm}
\end{center}
\end{table}

\section{Conclusions and Broader Impact}\label{sec:concl}

In this paper, we propose a spiking version of a 3-D and  hybrid combination of 3-D and 2-D convolutional architectures for HSI classification. We present a quantization-aware training technique, that yields highly accurate low-precision SNNs, which can be accelerated by integer math units or PIM accelerators. Our quantized SNNs offer significant improvements in energy consumption compared to both full and low-precision ANNs for HSI classification.

Our proposal results in energy-efficient SNN models, which can be readily deployed in HSI sensors, thereby eliminating the bandwidth and privacy concerns of going to the cloud. Since the commercial applications of HSI analysis are broadly expanding and the models required to train HSI are becoming deeper, energy-efficiency becomes a key concern, as seen in traditional computer vision tasks. To the best of our knowledge, this work is the first to address energy-efficiency of HSI models, and can hopefully inspire more research in low power algorithm-hardware co-design of neural network models for HSI classification.


%

\ifCLASSOPTIONcaptionsoff
  \newpage
\fi




{\small
\bibliographystyle{IEEEtran}
\bibliography{biblio}
}

%



%

\begin{IEEEbiography}[{\includegraphics[width=1in,height=1.25in,clip,keepaspectratio]{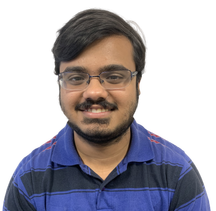}}]
{Gourav Datta} received his bachelors' degree in Instrumentation Engineering with a minor in Electronics and Electrical Communication Engineering from Indian Institute of Technology (IIT) Kharagpur, India in 2018. He then joined the Ming Hsieh Department of Electrical and Computer Engineering at the University of Southern California where he is currently pursuing a PhD degree. He interned at Apple Inc. and INRIA Research Centre in the summers of 2019 and 2017, respectively. His research focuses on the entire computing stack, including devices, circuits, architectures and algorithms for accelerating machine learning workloads. During his tenure at IIT Kharagpur, he has received the Institute Silver medal and was adjudged the best outgoing student in academics in his batch.
\end{IEEEbiography}
\vspace{-11mm}

\begin{IEEEbiography}[{\includegraphics[width=1in,height=1.25in,clip,keepaspectratio]{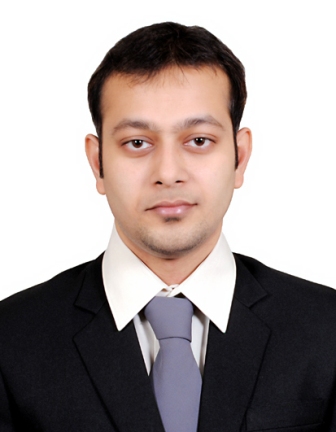}}]
{Souvik Kundu}
received his  M.\,Tech degree in Microelectronics and VLSI design from Indian Institute of Technology Kharagpur, India in 2015. He worked as R \& D Engineer II  at Synopsys India Pvt. Ltd. and as Digital Design Engineer at Texas Instruments India Pvt. Ltd. from 2015 to 2016 and from 2016 to 2017, respectively. He is currently working towards the Ph.D. degree in Electrical and Computer Engineering at the University of Southern California, Los Angeles, CA, USA. His research focuses on energy aware sparsity, model search, algorithm-hardware co-design of robust and energy-efficient neural networks for CMOS and beyond CMOS technology.
\end{IEEEbiography}
\vspace{-11mm}

\begin{IEEEbiography}[{\includegraphics[width=1in,height=1.25in,clip,keepaspectratio]{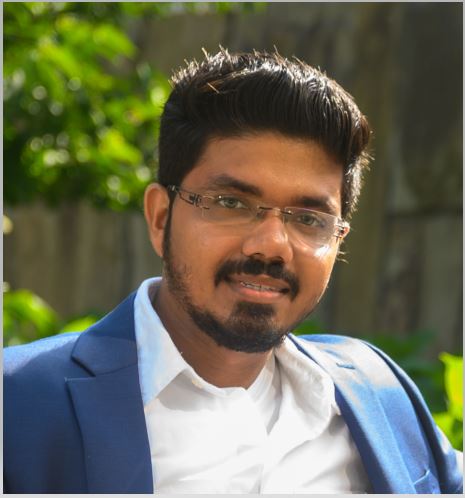}}]
{Akhilesh R.~Jaiswal} is a Research Assistant Professor of Electrical and Computer Engineering and a Scientist at USC's Information Sciences Institute's (ISI) Application Specific Intelligent Computing (ASIC) Lab. Prior to USC/ISI, Dr. Jaiswal was a Senior Research Engineer with GLOBALFOUNDIRES (GF) at Malta. Dr. Jaiswal received his Ph.D. degree in Nano-electronics from Purdue University in May 2019. As a part of doctoral program his research focused on 1) CMOS based analog and digital in-memory and near-memory computing using standard memory bit-cells for beyond von-Neumann computing. 2) Exploration of bio-mimetic devices and circuits using emerging non-volatile technologies for Neuromorphic computing. His current research interest includes exploration of 'alternate computing paradigms' using 'alternate state variables'. Dr. Jaiswal has authored several publications and holds 15+ issued and several pending patents with the USPTO.

\end{IEEEbiography}
\vspace{-11mm}

\begin{IEEEbiography}[{\includegraphics[width=1in,height=1.25in,clip,keepaspectratio]{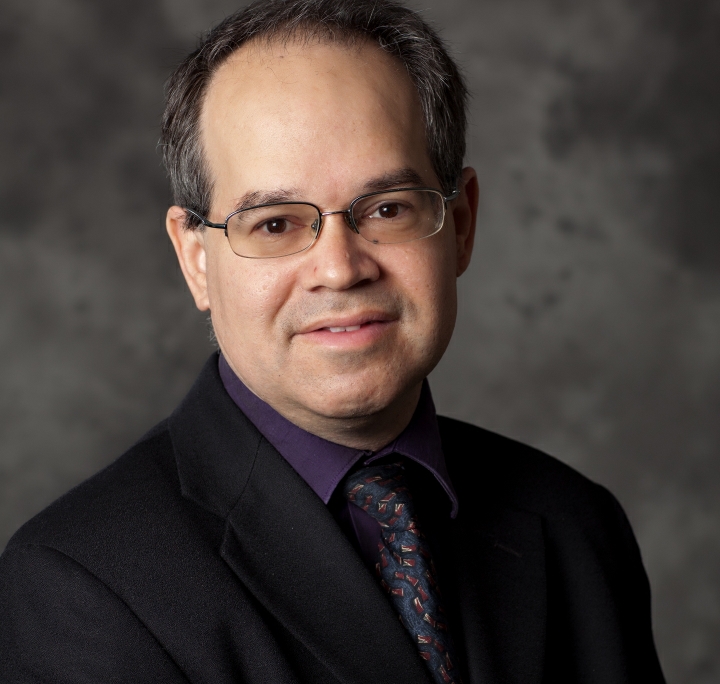}}]
{Peter A.~Beerel} received his B.S.E. degree in Electrical Engineering from Princeton University, Princeton, NJ, in 1989 and his M.S. and Ph.D. degrees in Electrical Engineering from Stanford University, Stanford, CA, in 1991 and 1994, respectively. He then joined the Ming Hsieh Department of Electrical and Computer Engineering at the University of Southern California where he is currently a professor and the Associate Chair of the Computer Engineering Division. He is also a Research Director at the Information Science Institute at USC. Previously, he co-founded TimeLess Design Automation to commercialize an asynchronous ASIC flow in 2008 and sold the company in 2010 to Fulcrum Microsystems which was bought by Intel in 2011. His interests include a variety of topics in computer-aided design, machine learning, hardware security, and asynchronous VLSI and the commercialization of these technologies. He is a Senior Member of the IEEE.
\end{IEEEbiography}




\end{document}